\documentclass[11pt,a4paper]{article}
\usepackage[hyperref]{acl2019}
\usepackage{times}
\usepackage{latexsym}
\usepackage{booktabs}
\usepackage{url}
\usepackage{graphicx}
\usepackage{amsmath}
\usepackage[ruled,linesnumbered]{algorithm2e}
\aclfinalcopy 

\makeatletter
\newcommand{\nosemic}{\renewcommand{\@endalgocfline}{\relax}}
\newcommand{\dosemic}{\renewcommand{\@endalgocfline}{\algocf@endline}}
\makeatother
\title{Soft Contextual Data Augmentation for Neural Machine Translation}

\author{Jinhua Zhu$^{1,*}$, Fei Gao$^{2,}$\thanks{\;\,The first two authors contributed equally to this work. This work is conducted at Microsoft Research Asia.}, Lijun Wu$^3$, Yingce Xia$^{4}$, \textbf{Tao Qin}$^4$, \\ \textbf{Wengang Zhou}$^1$, \textbf{Xueqi Cheng}$^2$, \textbf{Tie{-}Yan Liu}$^4$ \\
  $^1$University of Science and Technology of China, \\ $^2$Institute of Computing Technology, Chinese Academy of Sciences;\\ $^3$Sun Yat-sen University, $^4$Microsoft Reserach Asia;\\
  $^1${\texttt{\{teslazhu@mail., zhwg@\}ustc.edu.cn}},\\$^2$\texttt{\{gaofei17b, cxq\}@ict.ac.cn}, $^3$\texttt{wulijun3@mail2.sysu.edu.cn}, \\ $^4$\texttt{\{Yingce.Xia, taoqin, tyliu\}@microsoft.com}}
   
\date{}

\begin{document}
\def\aclpaperid{581}
\maketitle
\begin{abstract}
While data augmentation is an important trick to boost the accuracy of deep learning methods in computer vision tasks, its study in natural language tasks is still very limited. In this paper, we present a novel data augmentation method for neural machine translation. Different from previous augmentation methods that randomly drop, swap or replace words with other words in a sentence, we softly augment a randomly chosen word in a sentence by its contextual mixture of multiple related words. More accurately, we replace the one-hot representation of a word by a distribution (provided by a language model) over the vocabulary, i.e., replacing the embedding of this word by a weighted combination of multiple semantically similar words. Since the weights of those words depend on the contextual information of the word to be replaced, the newly generated sentences capture much richer information than previous augmentation methods. Experimental results on both small scale and large scale machine translation datasets demonstrate the superiority of our method over strong baselines\footnote{Our code can be found at \url{https://github.com/teslacool/SCA}}.

\end{abstract}

\section{Introduction}

Data augmentation is an important trick to boost the accuracy of deep learning methods by generating additional training samples. These methods have been widely used in many areas. For example, in computer vision, the training data are augmented by transformations like random rotation, resizing, mirroring and cropping \cite{krizhevsky2012imagenet, cubuk2018autoaugment}.


While similar random transformations have also been explored in natural language processing (NLP) tasks \cite{xie2017data}, data augmentation is still not a common practice in neural machine translation (NMT). For a sentence, existing methods include randomly swapping two words, dropping word, replacing word with another one and so on. However, due to text characteristics, these random transformations often result in significant changes in semantics.

A recent new method is contextual augmentation \cite{kobayashi2018contextual, wu2018conditional}, which replaces words with other words that are predicted using language model at the corresponding word position. While such method can keep semantics based on contextual information, this kind of augmentation still has one limitation: to generate new samples with adequate variation, it needs to sample multiple times. For example, given a sentence in which $N$ words are going to be replaced with other words predicted by one language model, there could be as many as exponential candidates. 
Given that the vocabulary size is usually large in languages, it is almost impossible to leverage all the possible candidates for achieving good performance.

In this work, we propose soft contextual data augmentation, a simple yet effective data augmentation approach for NMT. Different from the previous methods that randomly replace one word to another, we propose to augment NMT training data by replacing a randomly chosen word in a sentence with a \textit{soft word}, which is a probabilistic distribution over the vocabulary. Such a distributional representation can capture a mixture of multiple candidate words with adequate variations in augmented data. 
To ensure the distribution reserving similar semantics with original word, we calculate it based on the contextual information by using a language model, which is pretrained on the training corpus. 

To verify the effectiveness of our method, we conduct experiments on four machine translation tasks, including IWSLT$2014$ German to English,  Spanish to English, Hebrew to English and WMT$2014$ English to German translation tasks.  
In all tasks, the experimental results show that our method can obtain remarkable BLEU score improvement over the strong baselines.

\section{Related Work}
We introduce several related works about data augmentation for NMT.

\citet{artetxe2017unsupervised} and \citet{lample2017unsupervised} randomly shuffle (swap) the words in a sentence, with constraint that the words will not be shuffled further than a fixed small window size. \citet{iyyer2015deep} and \citet{lample2017unsupervised} randomly drop some words in the source sentence for learning an autoencoder to help train the unsupervised NMT model. In \citet{xie2017data}, they replace the word with a placeholder token or a word sampled from the frequency distribution of vocabulary, showing that data noising is an effective regularizer for NMT. \citet{fadaee2017data} propose to replace a common word by low-frequency word in the target sentence, and change its corresponding word in the source sentence to improve translation quality of rare words. Most recently, \citet{kobayashi2018contextual} propose an approach to use the prior knowledge from a bi-directional language model to replace a word token in the sentence. Our work differs from their work that we use a soft distribution to replace the word representation instead of a word token. 


\section{Method}
In this section, we present our method in details. 

\subsection{Background and Motivations}

Given a source and target sentence pair $(s, t)$ where $s = (s_1, s_2, ..., s_T)$ and $t = (t_1, t_2, ..., t_{T'})$, a neural machine translation system models the conditional probability $p(t_1,...,t_{T'}\vert s_1,...,s_T)$. NMT systems are usually based on an encoder-decoder framework with an attention mechanism \cite{sutskever2014sequence, bahdanau2014neural}. 
In general, the encoder first transforms the input sentence with words/tokens $s_1,s_2,...,s_T$ into a sequence of hidden states $\{h_t\}_{t=1}^{T}$, and then the decoder takes the hidden states from the encoder as input to predict the conditional distribution of each target word/token $p(t_\tau \vert h_t, t_{<\tau} )$ given the previous ground truth target word/tokens. Similar to the NMT decoder, a language model is intended to predict the next word distribution given preceding words, but without another sentence as a conditional input. 
In NMT, as well as other NLP tasks, each word is assigned with a unique ID, and thus represented as an one-hot vector. For example, the $i$-th word in the vocabulary (with size $|V|$) is represented as a $|V|$-dimensional vector $(0, 0, ..., 1, ..., 0)$, whose $i$-th dimension is $1$ and all the other dimensions are $0$. 

Existing augmentation methods generate new training samples by replacing one word in the original sentences with another word \cite{wang2018switchout, kobayashi2018contextual, xie2017data, fadaee2017data}. However, due to the sparse nature of words, it is almost impossible for those methods to leverage all possible augmented data.
First, given that the vocabulary is usually large, one word usually has multiple semantically related words as replacement candidates. 
Second, for a sentence, one needs to replace multiple words instead of a single word, making the number of possible sentences after augmentation increases exponentially.  
Therefore, these methods often need to augment one sentence multiple times and each time replace a different subset of words in the original sentence with different candidate words in the vocabulary; even doing so they still cannot guarantee adequate variations of augmented sentences. 
This motivates us to augment training data in a \textit{soft} way.

\subsection{Soft Contextual Data Augmentation}

Inspired by the above intuition, we propose to augment NMT training data by replacing a randomly chosen word in a sentence with a \textit{soft word}. Different from the discrete nature of words and their one-hot representations in NLP tasks, we define a soft word as a  distribution over the vocabulary of $|V|$ words. That is, for any word $w \in V$, its soft version is $P(w) = (p_1(w), p_2(w), ..., p_{|V|}(w))$, where $p_j(w) \ge 0$ and $\sum_{j=1}^{|V|}p_j(w) = 1$.

Since $P(w)$ is a distribution over the vocabulary, one can sample a word with respect to this distribution to replace the original word $w$, as done in \citet{kobayashi2018contextual}. Different from this method, we directly use this distribution vector to replace a randomly chosen word from the original sentence. Suppose $E$ is the embedding matrix of all the $|V|$ words. The embedding of the soft word $w$ is 
\begin{equation}
    e_{w} = P(w)E = \sum_{j=0}^{\vert V \vert}p_{j}(w) E_j,
\end{equation}
which is the expectation of word embeddings over the distribution defined by the soft word.

The distribution vector $P(w)$ of a word $w$ can be calculated in multiple ways. In this work, we leverage a pretrained language model to compute $P(w)$ and condition on all the words preceding $w$. That is, for the $t$-th word $x_t$ in a sentence, we have 
$$p_j(x_t)=LM(w_j|x_{<t}),$$
where $LM(w_j|x_{<t})$ denotes the probability of the $j$-th word in the vocabulary appearing after the sequence $x_1, x_2, \cdots, x_{t-1}$.
Note that the language model is pretrained using the same training corpus of the NMT model. Thus the distribution $P(w)$ calculated by the language model can be regarded as a smooth approximation of the original one-hot representation, which is very different from previous augmentation methods such as random swapping or replacement. Although this distributional vector is noisy, the noise is aligned with the training corpus.

\begin{figure}[!htb]
    \centering
    \includegraphics[width=1.05\linewidth]{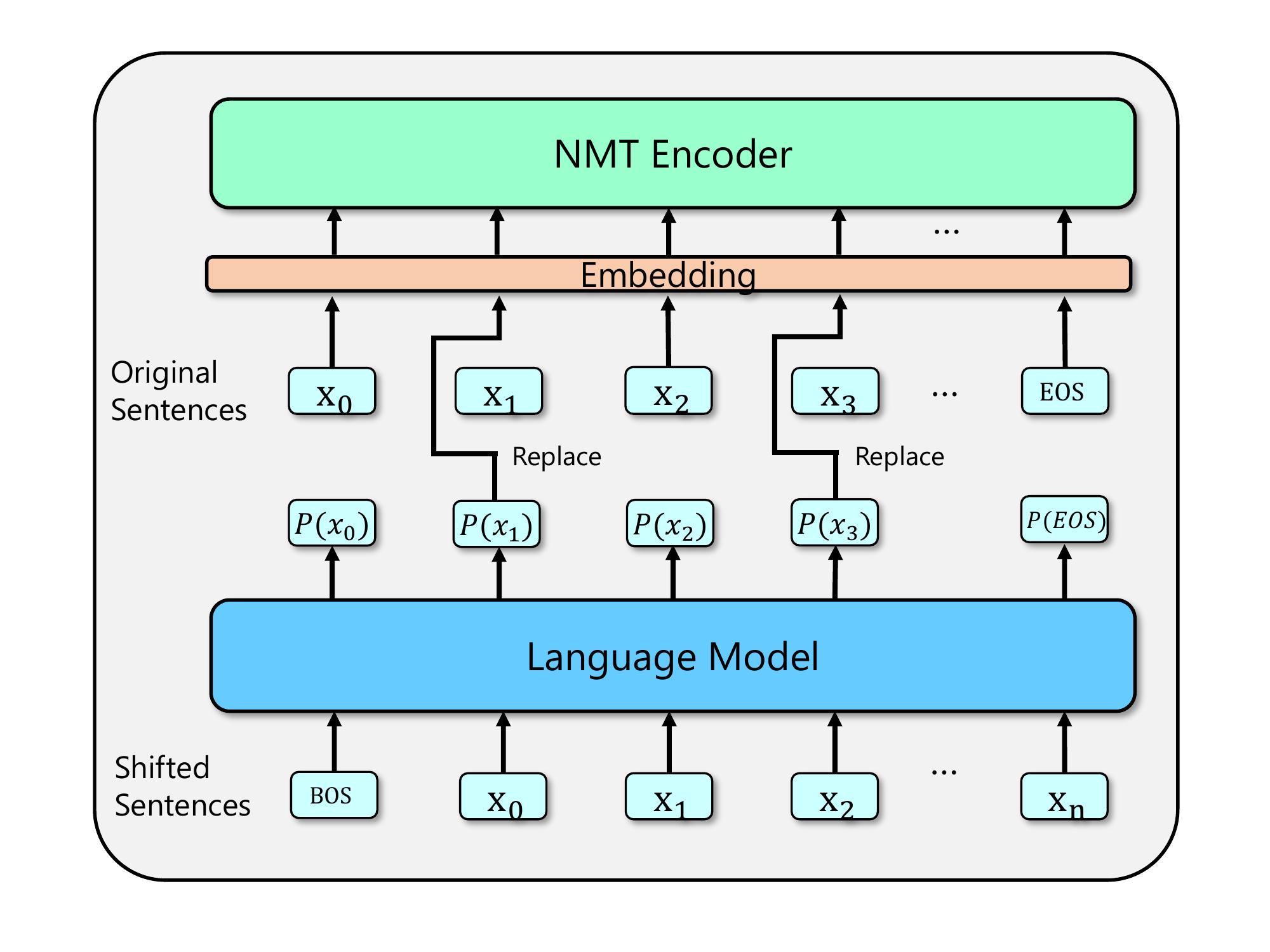}
    \caption{The overall architecture of our soft contextual data augmentation approach in encoder side for source sentences. The decoder side for target sentences is similar. }
    \label{fig:fig1}
\end{figure}

Figure \ref{fig:fig1} shows the architecture of the combination of the encoder of the NMT model and the language model. The decoder of the NMT model is similarly combined with the language model. In experiments, we randomly choose a word in the training data with probability $\gamma$ and replace it by its soft version (probability distribution). 

At last, it is worth pointing out that no additional monolingual data is used in our method. This is different from previous techniques, such as back translation, that rely on monolingual data \cite{sennrich2015improving, gulcehre2015using, cheng2016semi, he2016dual, hoang2018iterative}. We leave the exploration of leveraging monolingual data to future work.



\section{Experiment}\label{exp}
\begin{table*}[!htb]
\centering
\begin{tabular}{l c  c  c c  c}
\toprule
& \multicolumn{3}{c}{\textbf{IWSLT}} &&\textbf{WMT}\\
\cmidrule{2-4} \cmidrule{6-6}
 & \textbf{ De $\rightarrow$ En }  & \textbf{ Es $\rightarrow$ En } & \textbf{ He $\rightarrow$ En }&\ \ \   & \textbf{ En $\rightarrow$ De }  \\
\cmidrule{2-4} \cmidrule{6-6}
\emph{Base} & 34.79&41.58 &33.64 &&28.40 \\
\midrule
+\emph{Swap} &34.70 & 41.60 & 34.25 &&28.13\\
+\emph{Dropout} & 35.13&41.62 &34.29  &&28.29\\
+\emph{Blank}  &35.37 &42.28 & 34.37 &&28.89\\
+\emph{Smooth} &35.45 & 41.69& 34.61&& 28.97\\
+$LM_{sample}$ &35.40&42.09&34.31&&28.73\\
\midrule
\textbf{Ours} & \textbf{35.78} & \textbf{42.61}& \textbf{34.91}&&\textbf{29.70}\\
\bottomrule
\end{tabular}
\caption{BLEU scores on four translation tasks.}
\label{tab:iwslt_results}
\end{table*}
In this section, we demonstrate the effectiveness of our method on four translation datasets with different scale. The translation quality is evaluated by case-sensitive BLEU score. We compare our approach with following baselines: 
\begin{itemize}
    \item \emph{Base}: The original training strategy without any data augmentation;
    \item \emph{Swap}: Randomly swap words in nearby positions within a window size $k$ \citep{artetxe2017unsupervised, lample2017unsupervised};
    \item \emph{Dropout:} Randomly drop word tokens \citep{iyyer2015deep, lample2017unsupervised};
    \item \emph{Blank}: Randomly replace word tokens with a placeholder token \citep{xie2017data};
    \item \emph{Smooth}: Randomly replace word tokens with a sample from the unigram frequency distribution over the vocabulary \citep{xie2017data};
    \item \emph{$LM_{sample}$}: Randomly replace word tokens sampled from the output distribution of one language model \citep{kobayashi2018contextual}.
\end{itemize}

All above introduced methods except \emph{Swap} incorporate a hyper-parameter, the probability $\gamma$ of each word token to be replaced in training phase. 
We set $\gamma$ with different values in $\{0, 0.05, 0.1, 0.15, 0.2\}$, and report the best result for each method. As for \emph{swap}, we use 3 as window size following \citet{lample2017unsupervised}.

For our proposed method, we train two language models for each translation task. One for source language, and the other one for target language. The training data for the language models is the corresponding source/target data from the bilingual translation dataset. 

\subsection{Datasets}
We conduct experiments on IWSLT$2014$ \{German, Spanish, Hebrew\} to English (\{De, Es, He\}$\rightarrow$En) and WMT$2014$ English to German (En$\rightarrow$De) translation tasks to verify our approach. 
We follow the same setup in \citet{gehring2017convs2s} for IWSLT$2014$ De$\rightarrow$En task. The training data and validation data consist of $160k$ and $7k$ sentence pairs. \emph{tst2010}, \emph{tst2011}, \emph{tst2012}, \emph{dev2010} and \emph{dev2012} are concatenated as our test data. For Es$\rightarrow$En and He$\rightarrow$En tasks, there are $181k$ and $151k$ parallel sentence pairs in each training set, and we use \emph{tst2013} as the validation set, \emph{tst2014} as the test set. For all IWSLT translation tasks, we use a joint source and target vocabulary with $10K$ byte-pair-encoding (BPE) \citep{sennrich2015neural} types. For WMT$2014$ En$\rightarrow$De translation, again, we follow \citet{gehring2017convs2s} to filter out $4.5M$ sentence pairs for training. We concatenate \emph{newstest2012} and \emph{newstest2013} as the validation set and use \emph{newstest2014} as test set. The vocabulary is built upon the BPE with $40k$ sub-word types.


\subsection{Model Architecture and Optimization}
We adopt the sate-of-the-art Transformer architecture \citep{vaswani2017attention} for language models and NMT models in our experiments. For IWSLT tasks, we take the $transformer\_base$ configuration, except a) the dimension of the inner MLP layer is set as $1024$ instead of $2048$ and b) the number of attention heads is $4$ rather than $8$. As for the WMT En$\rightarrow$De task, we use the default $transformer\_big$ configuration for the NMT model, but the language model is configured with $transformer\_base$ setting in order to speed up the training procedure. All models are trained by Adam \citep{kingma2014adam} optimizer with default learning rate schedule as \citet{vaswani2017attention}. Note that after training the language models, the parameters of the language models are fixed while we train the NMT models.


\subsection{Main Results}

The evaluation results on  four translation tasks are presented in Table \ref{tab:iwslt_results}. As we can see, our method can consistently achieve more than $1.0$ BLEU score improvement over the strong Transformer base system for all tasks. Compared with other augmentation methods, we can find that 1) our method achieves the best results on all the translation tasks and 2) unlike other methods that may not be powerful in all tasks, our method universally works well regardless of the dataset. Specially, on the large scale WMT 2014 En$\rightarrow$De dataset, although this dataset already contains a large amount of parallel training sentence pairs, our method can still outperform the strong base system by $+1.3$ BLEU point and achieve $29.70$ BLEU score. These results clearly demonstrate the effectiveness of our approach.


\subsection{Study}
\begin{figure}[!htb]
    \centering
    \includegraphics[width=1.05\linewidth]{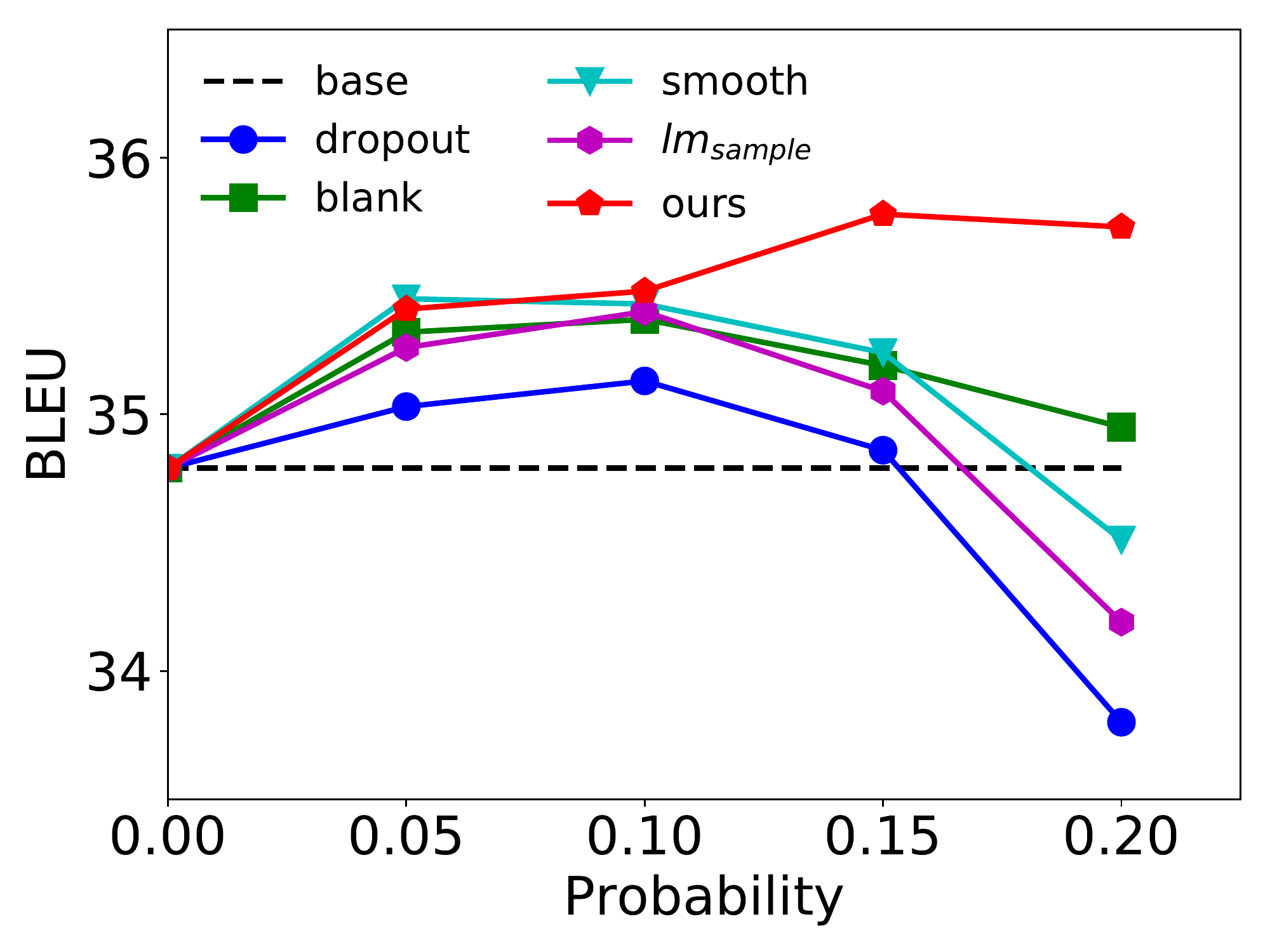}
    \caption{BLEU scores of each method on IWSLT De$\rightarrow$En dataset with different replacing probability.}
    \label{fig:fig2}
\end{figure}
As mentioned in Section \ref{exp}, we set different probability value of $\gamma$ to see the effect of our approach and other methods in this subsection.
Figure \ref{fig:fig2} shows the BLEU scores on IWSLT De$\rightarrow$En dataset of each method, from which we can see that our method can observe a consistent BLEU improvement within a large probability range and obtain a strongest performance when $\gamma = 0.15$. However, other methods are easy to lead to performance drop over the baseline if $\gamma > 0.15$, and the improvement is also limited for other settings of $\gamma$. This can again prove the superior performance of our method.                           

\section{Conclusions and Future Work}
In this work, we have presented soft contextual data augmentation for NMT, which replaces a randomly chosen word with a soft distributional representation. The representation is a probabilistic distribution over vocabulary and can be calculated based on the contextual information of the sentence. Results on four machine translation tasks have verified the effectiveness of our method.

In the future, besides focusing on the parallel bilingual corpus for the NMT training in this work, we are interested in exploring the application of our method on the monolingual data. In addition, we also plan to study our approach in other natural language tasks, such as text summarization. 


\bibliography{acl2019}
\bibliographystyle{acl_natbib}

\end{document}